\documentclass{article}

\usepackage[preprint]{corl_2026} 
\usepackage{graphicx}
\usepackage{algorithm}
\usepackage[utf8]{inputenc} 
\usepackage[T1]{fontenc}    
\usepackage{hyperref}       
\usepackage{url}            
\usepackage{booktabs}       
\usepackage{amsfonts}       
\usepackage{nicefrac}       
\usepackage{microtype}      
\usepackage{multicol}
\usepackage{multirow}
\usepackage{amsmath}
\usepackage{wrapfig}
\usepackage{subcaption}
\usepackage{caption}
\usepackage{booktabs}
\usepackage{adjustbox}
\usepackage{url}
\usepackage{placeins}
\usepackage{pifont}

\newcommand{\cmark}{\ding{51}}
\newcommand{\xmark}{\ding{55}}

\title{Uncovering Vulnerability of Vision-Language-Action Models under Joint-Level Physical Faults}

\author{
  Minsoo Jo$^1$\footnotemark[1] \: Taeju Kwon$^1$\thanks{Equal Contribution} \: Junha Chun$^1$ \: Youngjoon Jeong$^1$ \: Taesup Kim$^1$\thanks{Corresponding Author} \: \\
  $^1$ Graduate School of Data Science, Seoul National University\\
}

\begin{document}

\maketitle
\begin{abstract}
Deploying Vision-Language-Action (VLA) models in real robotic systems requires robustness not only to semantic and perceptual variations, but also to embodiment-side faults that change how actions are physically realized. 
Real robots can experience joint-level changes caused by actuator degradation, hardware faults, safety limits, collision damage, or wear-induced friction. 
These faults are critical because they alter the action-to-motion interface of a policy, disrupting the learned closed-loop relationship between commanded actions, realized motion, and subsequent observations.
In this work, we study realistic joint-level physical faults and show that VLA models are vulnerable when predicted actions are executed through a perturbed robot body.
Our analysis reveals joint-dependent effects, with heterogeneous degradation in task success across affected joints.
We also show that performance drops cannot be attributed solely to physical infeasibility, since feasible faults such as increased joint friction can still substantially reduce success rates and induce closed-loop execution mismatch.
Motivated by these findings, we propose Joint-level Physical-fault Aware Residual Calibrator (J-PARC), a lightweight residual calibration framework built on top of a frozen VLA policy. 
J-PARC infers a latent joint-fault regime from recent joint dynamics and conditions a shared residual calibrator on this regime, enabling adaptive action correction across faulty joints. 
Experiments show that J-PARC improves robustness under joint-level faults while preserving fault-free environment performance.

\end{abstract}
\section{Introduction}


Deploying Vision-Language-Action (VLA) models~\cite{openpi05, openpi, univla, openvla, openvlaoft} in real robotic systems requires robustness not only to perception-side disturbances~\cite{psd1,psd2,liberoplus}, but also to embodiment-side faults that affect how actions are physically realized.
While recent VLA models have shown strong generalization across tasks and visual conditions, their deployment often assumes that the robot body faithfully converts policy-predicted actions into intended motions.
This assumption can break down in real robots, where actuator degradation, partial hardware faults, safety-imposed joint limits, collision-induced damage, mechanical wear, or lubricant degradation can change the robot’s joint-level execution capability.
Because such faults may persist during autonomous operation and cannot always be immediately detected, repaired, or reset, a deployed robot must remain reliable even when its physical execution capability is partially degraded.
\textit{This makes embodiment-side robustness a necessary condition for reliable VLA deployment, rather than a secondary hardware concern.}

Joint-level physical faults are particularly important because they directly alter the action-to-motion interface through which a VLA policy interacts with the world.
Even when the policy predicts a reasonable action, the robot body may realize a different motion under a faulted joint.
Repeated over multiple control steps, this discrepancy can cause the robot to drift from the intended task trajectory and potentially lead to unsafe or unrecoverable failures~\cite{joint_fault1,joint_fault2,joint_fault3}.
In this work, we study joint-level physical faults including joint lock, limited range of motion~\cite{lock,lock2,lock3}, and increased joint friction~\cite{friction,friction2,friction3}.
We treat range limits and friction as gradual and realistic degradation modes that generalize the fully locked-joint case by progressively restricting or distorting joint mobility.

\begin{figure*}[t]
    \centering
    \includegraphics[width=\textwidth]{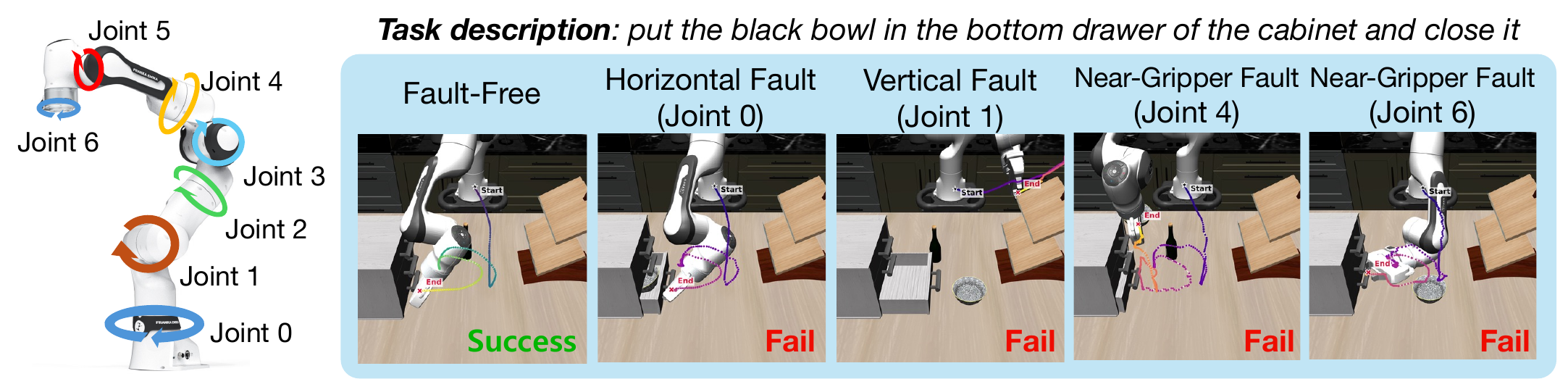}
    \vspace{-1.5em}
    \caption{
    \textbf{VLA models under joint-level physical faults.}
    A VLA policy successfully completes the LIBERO~\cite{libero} task in the fault-free setting, but fails when different Franka Panda joints are locked.
    This illustrates how embodiment-side faults can change the robot's realized motion without changing the policy output, motivating physical-fault aware action calibration.
    }
    \vspace{-1.5em}
    \label{fig:vla_fault}
\end{figure*}

We show that VLA models are vulnerable to these faults when their predicted actions are executed through a faulted robot body.
As illustrated in Fig.~\ref{fig:vla_fault}, the same VLA policy that succeeds with a fault-free robot body can fail when individual joints are physically constrained.
This observation motivates the need to understand and compensate for the mismatch between policy-commanded actions and robot-realized motions.

Joint-level physical faults introduce a robustness challenge that is qualitatively different from conventional action noise.
Command-space perturbations directly modify the policy output or action command, whereas joint-level physical faults alter the robot’s kinematic or dynamic capability itself.
As a result, the same commanded action can lead to different realized states depending on the affected joint, fault type, and fault severity.
\textit{Thus, robustness to perturbed action commands does not necessarily imply robustness to a perturbed robot body.}
This mismatch is not merely a single-step execution error.
Subsequent observations deviate from the nominal rollout, causing the policy to act on off-nominal states.
As this closed-loop execution mismatch accumulates over time, the model may produce actions that become increasingly misaligned with task completion~\cite{ood}.

Our analysis shows that the effects of joint-level faults are joint-dependent, leading to heterogeneous degradation in task success across affected joints.
Moreover, the observed performance drop cannot be attributed solely to physical infeasibility, since feasible faults such as increased joint friction can still substantially reduce success rates.
These findings indicate that robust VLA deployment requires mechanisms that can infer and compensate for latent embodiment faults during execution.

To address this problem, we propose Joint-level Physical fault-Aware Residual Calibrator (J-PARC), a lightweight residual calibration framework built on top of a frozen VLA policy.
J-PARC estimates the current joint-fault regime from recent robot joint dynamics and uses this regime as context for a shared residual calibrator that predicts arm-action corrections.
By adapting the commanded action to the observed execution context, J-PARC compensates for the mismatch between policy-commanded actions and robot-realized motions without fine-tuning the base VLA policy or repeatedly generating adversarial samples.
This design allows the base policy to retain its behavior under fault-free conditions while adding an execution-aware correction layer for faulted embodiments.

Our main contributions are as follows:
\begin{itemize}
\item We identify joint-level physical faults as a critical embodiment-side robustness problem for VLA deployment and analyze how locked joints, limited range of motion, and increased friction affect embodied robot execution.
\item We show that joint-level faults induce joint-dependent task degradation and closed-loop execution mismatch that cannot be fully explained by physical infeasibility alone.
\item We propose J-PARC, a lightweight residual calibration framework that infers joint-fault regimes from recent joint dynamics and improves robustness across different joints and fault modes while preserving the behavior of the frozen VLA policy under fault-free conditions.
\end{itemize}

\begin{figure}[t]
    \centering
    \captionsetup[subfigure]{justification=centering, singlelinecheck=false}
    \begin{subfigure}[t]{0.62\textwidth}
        \centering
        \includegraphics[width=\linewidth]{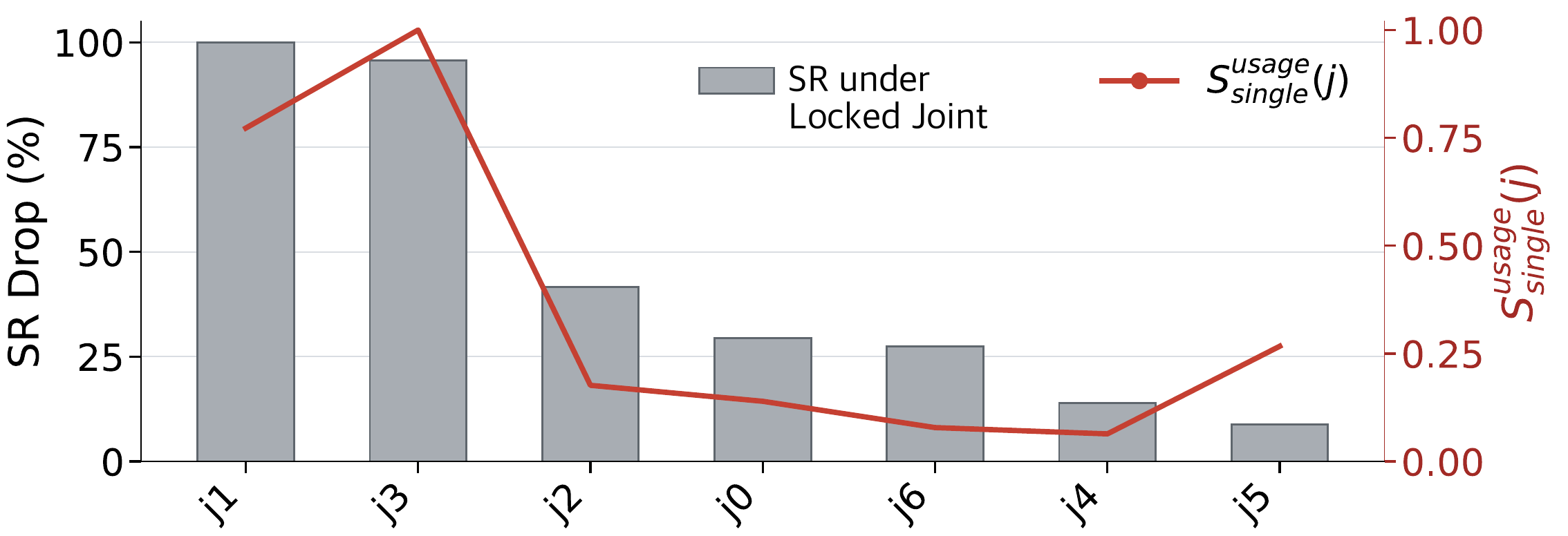}
        \caption{Success rate under locked-joint faults and weighted EEF sensitivity.}
        \label{fig:joint_fault_heterogeneity}
    \end{subfigure}
    \hfill
    \begin{subfigure}[t]{0.37\textwidth}
        \centering
        \includegraphics[width=\linewidth]{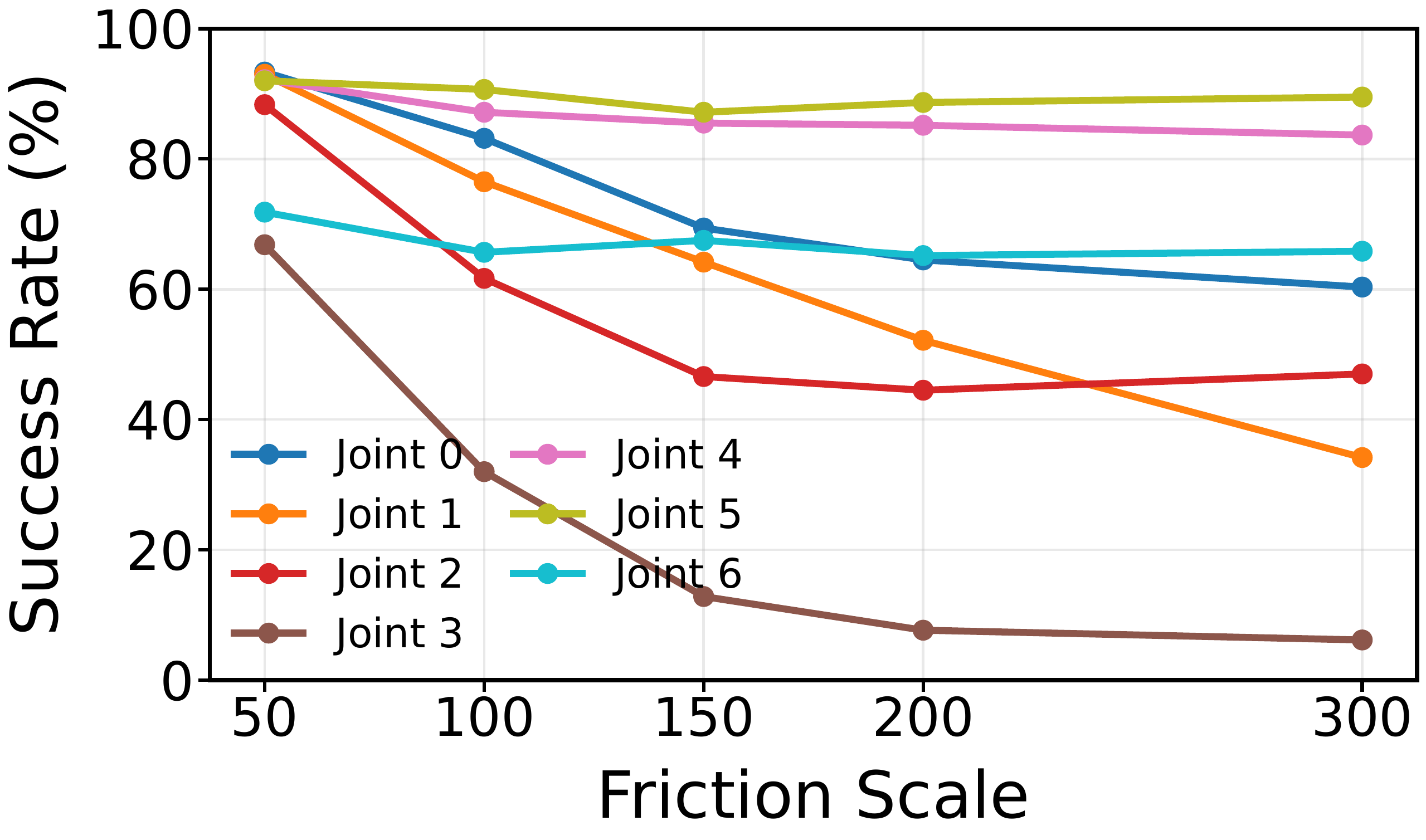}
        \caption{Success rate under friction faults.}
        \label{fig:friction_feasible}
    \end{subfigure}

    \caption{
    \textbf{Heterogeneous and feasible joint-level fault effects.}
    Joint-level faults cause heterogeneous performance degradation across joints, and increased friction can reduce success rates even when the task remains physically feasible.
    }
    \label{fig:joint_fault_combined}
    \vspace{-1.0em}
\end{figure}

\section{VLA Models are Vulnerable to Joint-Level Physical faults}
We first investigate the vulnerability of VLA policies to joint-level physical faults during closed-loop robot execution. 
Through joint lock, range limitation, and friction faults, we show that VLA performance degradation is joint-dependent, cannot be fully attributed to physical infeasibility, and is amplified by closed-loop execution mismatch. 
\subsection{Heterogeneous Effects of Joint-Level Faults}

Fig.~\ref{fig:vla_fault} further illustrates that faults on Joints 0 mainly restrict horizontal motion, while faults on Joints 4 and 6, which are closer to the end effector, interfere with grasping and object handling.
Faults on Joints 1 primarily affect vertical motion and often render many tasks infeasible.

We compute the usage-weighted end-effector sensitivity of each joint from fault-free rollouts by multiplying its task-specific joint-range usage by the rollout-averaged maximum translational Jacobian norm of gripper probe sites with respect to that joint, with details provided in Appendix~\ref{app:sensitivity}. As shown in Fig.~\ref{fig:joint_fault_heterogeneity}, the effect of joint-level fault conditions on task performance is highly heterogeneous.
This heterogeneity arises from differences in both joint usage and the sensitivity of end-effector motion to each joint.
Moreover, as shown in Fig.~\ref{fig:friction_feasible}, increasing the joint-friction scale reduces the success rate even when the tasks remain feasible under the friction fault setting.
Unlike joint lock or range-limit faults, increased friction does not remove task-relevant reachable configurations; instead, it alters how commanded actions are realized through the robot dynamics.
This result suggests that VLA performance degradation is also caused by execution mismatch under altered joint dynamics, not only by unreachable task states.
These observations indicate that motion correction should be conditioned on which joint is under a fault condition, as considered in Sec.~\ref{sec:fault_conditioned_training}.

\subsection{Fault Accumulation Reveals Limited Recovery from Closed-Loop Mismatch}
\begin{wrapfigure}{r}{0.31\textwidth}
    \vspace{-1.15em}
    \includegraphics[width=\linewidth]{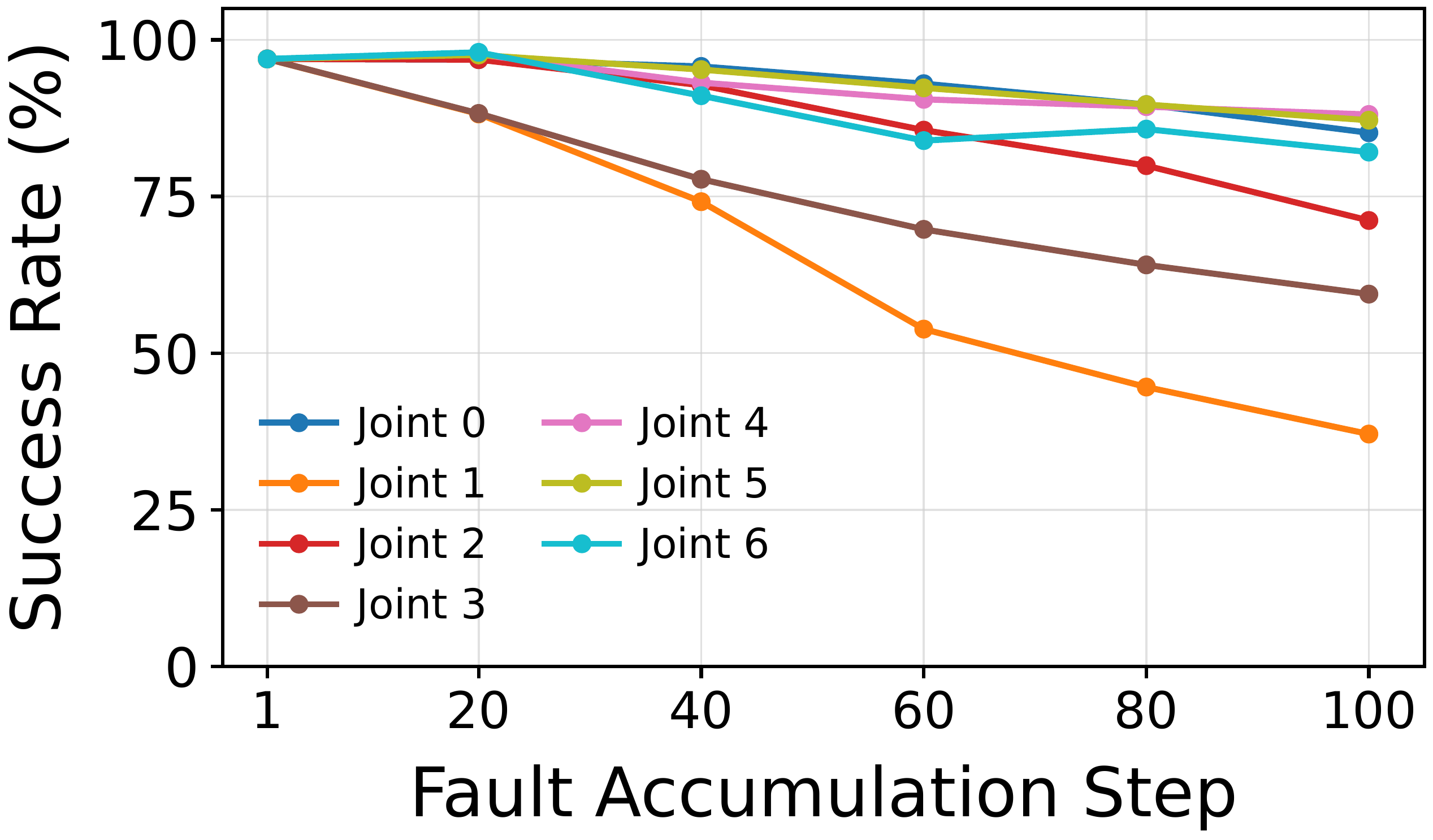}
    \vspace{-1.25em}
    \caption{
    \textbf{Fault-accumulation recovery.}
    Success drops as faults persist before release.
    }
    \vspace{-1.25em}
    \label{fig:fault_accumulation_recovery}
\end{wrapfigure}
To examine whether VLA policies can recover from states induced by persistent joint faults, we evaluate a fault-accumulation setting.
The robot first executes under a locked-joint fault for a specified number of steps, allowing the fault-induced deviation to accumulate in the robot state.
Then, starting from this accumulated state, the policy is evaluated under fault-free dynamics to test whether it can still complete the task.
As shown in Fig.~\ref{fig:fault_accumulation_recovery}, task success decreases as the fault is accumulated for more steps.
This indicates that joint faults do not merely cause transient execution errors; they progressively move the robot away from the nominal state trajectory expected by the VLA policy.
Once the accumulated state lies far from this in-distribution rollout, even fault-free execution may be insufficient for recovery.
Therefore, overcoming joint-level faults requires calibrating actions during the fault condition so that the robot state remains close to the in-distribution trajectory of the base policy, as addressed by the reference-guided residual targets in Sec.~\ref{sec:reference_guided_targets}.

\subsection{Existing Robustness Methods Do Not Fully Capture Joint-Level Fault Vulnerabilities}

\begin{wrapfigure}{r}{0.50\textwidth}
    \vspace{-1.0em}
    \centering
    \captionsetup{singlelinecheck=false}

    \newlength{\panelheight}
    \setlength{\panelheight}{0.20\textheight}

    \begin{minipage}[t]{0.54\linewidth}
        \vspace{0pt}
        \centering
        \begin{minipage}[t][\panelheight][t]{\linewidth}
            \vspace{0pt}
            \centering
            \includegraphics[
                width=\linewidth,
                height=\panelheight,
                keepaspectratio
            ]{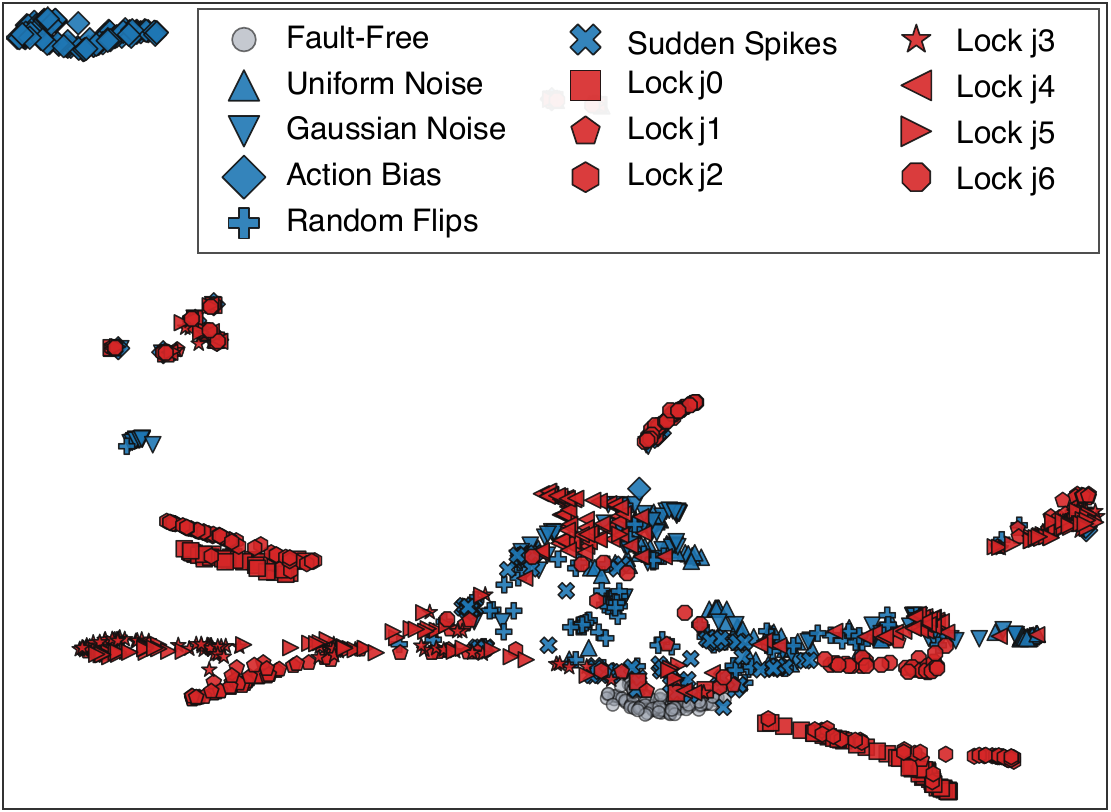}
        \end{minipage}
        \vspace{-4.6em}
        \captionof{figure}{
        UMAP visualization of robot state transition distributions under different fault conditions.
        }
        \label{fig:robot_state_umap_stuck_noise}
    \end{minipage}
    \hfill
    \begin{minipage}[t]{0.38\linewidth}
        \vspace{0pt}
        \centering
        \begin{minipage}[t][\panelheight][t]{\linewidth}
            \vspace{0pt}
            \centering
            \setlength{\tabcolsep}{3pt}
            \renewcommand{\arraystretch}{0.85}

            \begin{adjustbox}{
                height=\panelheight,
                max width=\linewidth,
                keepaspectratio
            }
            \begin{tabular}{lccc}
            \toprule
            Joint & $\pi_{0.5}$ & CIK & $\Delta$ \\
            \midrule
            $j_0$ & 57.4 & 54.7 & -2.6 \\
            $j_1$ & 0.0  & 0.0  & +0.0 \\
            $j_2$ & 37.5 & 33.3 & -4.2 \\
            $j_3$ & 6.3  & 5.4  & -0.9 \\
            $j_4$ & 85.2 & 83.5 & -1.7 \\
            $j_5$ & 86.9 & 82.2 & -4.8 \\
            $j_6$ & 66.7 & 68.8 & +2.1 \\
            \midrule
            Mean & 48.6 & 46.8 & -1.7 \\
            \bottomrule
            \end{tabular}
            \end{adjustbox}
        \end{minipage}
        \vspace{-4.6em}
        \captionof{table}{
        CIK under locked-joint faults.
        }
        \label{tab:cik_stuck_table_panel}
    \end{minipage}

    \vspace{-1.0em}
\end{wrapfigure}
Fig.~\ref{fig:robot_state_umap_stuck_noise} shows that robot states induced by action noise differ significantly from those induced by locked-joint fault conditions. While action noise perturbs the policy output directly, joint-level faults change how actions are physically realized through the robot body, producing distinct state distributions.
This suggests that command-space robustness, as considered in prior methods such as RobustVLA~\cite{robustvla}, does not necessarily prepare VLA models for persistent embodiment-side faults.

Another natural baseline is Constrained Inverse Kinematics (CIK)~\cite{cik1, cik2}, which uses identified joint constraints to move the end effector closer to the target pose implied by the VLA action.
To isolate the effect of CIK, Tab.~\ref{tab:cik_stuck_table_panel} reports CIK results under an oracle assumption where the faulty joint is perfectly identified.
Despite this favorable setting, CIK does not consistently improve over the base $\pi_{0.5}$ policy and even reduces the mean success rate.
This is because CIK uses the identified faulty joint only to impose local constraints and performs per-step correction toward the target pose, without access to rollout history or the in-distribution semantics expected by the VLA policy. Thus, even with oracle fault identification, CIK can reduce local kinematic error but cannot prevent closed-loop execution mismatch, as joint reconfiguration and small residual errors can gradually drive the trajectory away from the nominal distribution.

These observations show that joint-level faults require more than command-space distribution expansion or local kinematic correction.
Instead, the policy must infer the current execution context from recent robot states and VLA actions and calibrate actions to keep the rollout under fault condition close to the base policy's learned trajectory distribution, as formalized in Eqs.~\eqref{eq:history_buffer} and~\eqref{eq:corrected_action}. 
\section{Method}
\label{sec:method}

We propose \emph{Joint-level Physical-fault Aware Residual Calibrator} (J-PARC), a lightweight module attached to a frozen Vision-Language-Action (VLA) policy. J-PARC observes recent joint dynamics and execution history, estimates the current fault context, and adds a residual only to the translational arm command.

\subsection{J-PARC Overview}
\label{sec:jparc_overview}

Let $\pi_\phi$ be a pretrained VLA policy. Given visual-proprioceptive observation $o_t$ and language instruction $\ell$, it predicts an action chunk
\begin{align*}
    A_t^{\mathrm{base}}
    &= \pi_\phi(o_t,\ell)
    = \{a_{t,k}^{\mathrm{base}}\}_{k=0}^{K-1}
    \in\mathbb{R}^{K\times 7}, \\
    a_{t,k}^{\mathrm{base}} 
    &= [\Delta x_{t,k}^{\mathrm{base}},\Delta y_{t,k}^{\mathrm{base}},\Delta z_{t,k}^{\mathrm{base}},
    \Delta r_{x,t,k}^{\mathrm{base}},\Delta r_{y,t,k}^{\mathrm{base}},\Delta r_{z,t,k}^{\mathrm{base}},g_{t,k}^{\mathrm{base}}] \\
    &=
    [a_{t,k}^{\mathrm{base},xyz},
     a_{t,k}^{\mathrm{base},rot},
     a_{t,k}^{\mathrm{base},g}],
    \label{eq:base_action_chunk}
\end{align*}
where $K$ is the action-chunk horizon. The observation contains visual, task, end-effector, gripper, and arm-joint states.

J-PARC maintains a step-wise execution history
\begin{equation}
    \mathcal{H}_t=\{h_i\}_{i=t-H}^{t-1},\qquad
    h_i=[a_i^{\mathrm{base}}, c_i, \delta_i^{\mathrm{hist}}],
    \qquad
    c_i=[s_i,q_i,\dot q_i],
    \label{eq:history_buffer}
\end{equation}
where $H$ is the history length, $s_i$ denotes the end-effector and gripper state, and $\delta_i^{\mathrm{hist}}=a_i^{\mathrm{exec},xyz}-a_i^{\mathrm{base},xyz}$ records the difference between the base action and the action executed at step $i$.

Conditioned on a joint-fault latent $u_t$ in Eq.~\ref{eq:fault_encoder}, J-PARC predicts a residual chunk in parallel over the VLA action chunk:
\begin{equation}
    R_{\theta,t}
    =
    \{r_{\theta,t,k}\}_{k=0}^{K-1}
    =
    f_\theta(c_t,A_t^{\mathrm{base}},\mathcal{H}_t,u_t)
    \in\mathbb{R}^{K\times3}.
    \label{eq:jparc_residual}
\end{equation}
The same history format is used during training and deployment: $a_i^{\mathrm{exec}}$ is the logged behavior-policy action during offline training and J-PARC's previously corrected action during deployment.

For the $k$-th action in the chunk, the corrected action is
\begin{equation}
    \tilde a_{t,k}^{xyz}
    =
    a_{t,k}^{\mathrm{base},xyz}+r_{\theta,t,k},
    \qquad
    \tilde a_{t,k}^{rot}=a_{t,k}^{\mathrm{base},rot},
    \qquad
    \tilde a_{t,k}^{g}=a_{t,k}^{\mathrm{base},g}.
    \label{eq:corrected_action}
\end{equation}
J-PARC calibrates only the translational action, $a_{t,k}^{\mathrm{base},xyz}$, while preserving the frozen VLA's rotation and gripper commands, as orientation correction under joint faults can induce unnecessary joint reconfiguration. The corrected chunk is queued and executed one action at a time in the environment. After each environment step, the actually applied residual is appended to $\mathcal{H}_t$ and used for subsequent replanning.

\subsection{Reference-Guided Residual Targets}
\label{sec:reference_guided_targets}

We collect supervision by pairing a fault-free reference rollout and rollouts under a locked-joint condition from the same initial state. The fault-free rollout provides reference end-effector positions
\begin{equation}
    \tau^{\mathrm{ref}}
    =
    \{x_i^{\mathrm{ref}}\}_{i=0}^{T}.
\end{equation}
During faulty execution, for each valid chunk index $k$, let $i=t+k$. We measure deviation from the next fault-free waypoint:
\begin{equation}
    d_i=\|x_i-x_{i+1}^{\mathrm{ref}}\|_2.
    \label{eq:demo_deviation}
\end{equation}
If the cached reference is exhausted, a synchronized fault-free replan supplies the next reference waypoint.

The threshold $\tau_{\mathrm{demo}}$ defines a demonstration tube. Inside the tube, the teacher leaves the base action unchanged; outside it, the teacher redirects translation toward the fault-free reference:
\begin{equation}
    a_{t,k}^{\star,xyz}=
    \begin{cases}
    a_{t,k}^{\mathrm{base},xyz}, & d_i\leq\tau_{\mathrm{demo}},\\
    \alpha_i (x_{i+1}^{\mathrm{ref}}-x_i), & d_i>\tau_{\mathrm{demo}},
    \end{cases}
    \qquad
    \alpha_i=
    \frac{\|a_{t,k}^{\mathrm{base},xyz}\|_2}
    {\max(\|x_{i+1}^{\mathrm{ref}}-x_i^{\mathrm{ref}}\|_2,\epsilon)}.
    \label{eq:thresholded_teacher}
\end{equation}
Rotation and gripper commands are kept identical to the base VLA action. The residual target for each valid chunk element is
\begin{equation}
    r_{t,k}^{\star}
    =
    a_{t,k}^{\star,xyz}
    -
    a_{t,k}^{\mathrm{base},xyz}.
    \label{eq:residual_target}
\end{equation}
This makes the target zero near the fault-free trajectory and non-zero only when the fault produces meaningful closed-loop deviation.

\subsection{Fault-Conditioned Training}
\label{sec:fault_conditioned_training}

To condition on the fault regime without using fault labels at deployment, we encode recent joint motion as
\begin{equation}
    u_t=g_\psi(\Delta q_{t-H:t}),
    \qquad
    u_t\in\mathbb{R}^{d_u}.
    \label{eq:fault_encoder}
\end{equation}
The encoder $g_\psi$ is pretrained with two supervised objectives: a binary cross-entropy loss for detecting whether the rollout is under joint fault, and a cross-entropy loss for classifying the affected joint among perturbed rollouts. It is implemented as a temporal encoder with a per-step MLP followed by a GRU, and is frozen during residual training.

The residual calibrator $f_\theta$ is a transformer encoder over the current state token, step-wise history tokens, base action chunk tokens, and the fault latent token. We train only $f_\theta$, keeping $\pi_\phi$ and $g_\psi$ frozen, with
\begin{equation}
    \mathcal{L}_{\mathrm{res}}
    =
    \frac{1}{N}
    \sum_{t,k}
    \|r_{\theta,t,k}-r_{t,k}^{\star}\|_2^2,
    \label{eq:residual_loss}
\end{equation}
where $N$ is the number of supervised residual terms. During deployment, the base VLA remains frozen throughout.
\section{Experiments}
\label{sec:experiments}

\begin{table}[t]
\centering
\caption{
\textbf{LIBERO benchmark evaluation results under joint fault.}
Success rates (\%) averaged over LIBERO~\cite{libero} suites.
CIK$^*$ assumes oracle faulty-joint identification.
RVLA denotes RobustVLA~\cite{robustvla}, which is reproduced on $\pi_{0.5}$ by their official repository\protect\footnotemark.
CIK is omitted for friction faults because it cannot be evaluated in that setting.
}
\label{tab:main_results_single_joint_filtered}

\begingroup
\small
\setlength{\tabcolsep}{4.5pt}
\renewcommand{\arraystretch}{0.95}

\newlength{\jointtableheight}
\setlength{\jointtableheight}{0.37\textheight}

\vspace{0.3em}
\begin{minipage}[t]{0.54\textwidth}
\vspace{0pt}
\centering
\textbf{Joint limit}
\vspace{0.217em}

\begin{adjustbox}{width=\linewidth,totalheight=\jointtableheight}
\begin{tabular}{llccccccc}
\toprule
\multirow{2}{*}{\footnotesize {Range}} & \multirow{2}{*}{J.}
& \multicolumn{3}{c}{OpenVLA-OFT}
& \multicolumn{4}{c}{$\pi_{0.5}$} \\
\cmidrule(lr){3-5}\cmidrule(lr){6-9}
& & Base & CIK$^*$ & J-PARC
& Base & RVLA & CIK$^*$ & J-PARC \\
\midrule
\midrule
1.0 & - & \textbf{95.8} & \textbf{95.8} & 95.4 & 96.6 & 96.2 & 96.6 & \textbf{97.0}\\
\midrule

\multirow{8}{*}{0.05}
& $j_0$ & 81.5 & 76.8 & \textbf{90.6} & 85.5 & 82.6 & 83.2 & \textbf{93.8} \\
& $j_1$ & 0 & 0 & 0 & 0 & 0 & 0 & 0 \\
& $j_2$ & 70.6 & 65.2 & \textbf{77.4} & 81.6 & 81.5 & 75.7 & \textbf{90.4} \\
& $j_3$ & 9.8 & 7.6 & \textbf{10.2} & 11.8 & 11.2 & 9.9 & \textbf{12.4} \\
& $j_4$ & 84.9 & 80.8 & \textbf{85.7} & 87.0 & 87.4 & 85.0 & \textbf{88.5} \\
& $j_5$ & 83.2 & 75.8 & \textbf{84.5} & 91.7 & 89.5 & 87.0 & \textbf{93.4} \\
& $j_6$ & 65.0 & 64.0 & \textbf{68.9} & 71.4 & 70.0 & 72.3 & \textbf{74.6} \\
\cmidrule(lr){2-9}
& Avg. & 56.4 & 52.9 & \textbf{59.6} & 61.3 & 60.3 & 59 & \textbf{64.7} \\

\midrule

\multirow{8}{*}{0.03}
& $j_0$ & 72.2 & 67.8 & \textbf{80.2} & 75.5 & 75.1 & 74.7 & \textbf{87.5} \\
& $j_1$ & 0 & 0 & 0 & 0 & 0 & 0 & 0 \\
& $j_2$ & 56.3 & 52.5 & \textbf{61.4} & 66.3 & 66.0 & 61.0 & \textbf{78.7} \\
& $j_3$ & \textbf{8.2} & 5.9 & \textbf{8.2} & 8.0 & 7.3 & 7.2 & \textbf{8.7} \\
& $j_4$ & 83.5 & 79.6 & \textbf{84.0} & 87.0 & 86.3 & 85.0 & \textbf{87.8} \\
& $j_5$ & 81.5 & 72.9 & \textbf{83.8} & 90.5 & 87.9 & 84.7 & \textbf{92.2} \\
& $j_6$ & 62.9 & 62.6 & \textbf{66.9} & 69.0 & 69.4 & 70.0 & \textbf{73.8} \\
\cmidrule(lr){2-9}
& Avg. & 52.1 & 48.8 & \textbf{54.9} & 56.6 & 56.0 & 54.7 & \textbf{61.2} \\

\midrule

\multirow{8}{*}{Locked}
& $j_0$ & 48.8 & 46.5 & \textbf{59.6} & 57.4 & 55.3 & 54.7 & \textbf{72.2} \\
& $j_1$ & 0 & 0 & 0 & 0 & 0 & 0 & 0 \\
& $j_2$ & 27.4 & 22.5 & \textbf{34.2} & 37.5 & 36.0 & 33.3 & \textbf{56.1} \\
& $j_3$ & 4.7 & \textbf{4.8} & \textbf{4.8} & 5.1 & 4.8 & 5.4 & \textbf{5.4} \\
& $j_4$ & 79.0 & 75.3 & \textbf{80.5} & 85.2 & 83.7 & 83.5 & \textbf{87.6} \\
& $j_5$ & 78.2 & 67.5 & \textbf{83.2} & 86.9 & 86.7 & 82.2 & \textbf{91.2} \\
& $j_6$ & 58.2 & 59.0 & \textbf{65.1} & 66.7 & 66.0 & 68.8 & \textbf{73.8} \\
\cmidrule(lr){2-9}
& Avg. & 42.3 & 39.4 & \textbf{46.8} & 48.4 & 47.5 & 46.8 & \textbf{55.2} \\

\bottomrule
\end{tabular}
\end{adjustbox}
\end{minipage}
\hfill
\begin{minipage}[t]{0.42962\textwidth}
\vspace{0pt}
\centering
\textbf{Friction}
\vspace{0.3em}

\begin{adjustbox}{width=\linewidth,totalheight=\jointtableheight}
\begin{tabular}{llccccc}
\toprule
\multirow{2}{*}{Scale} & \multirow{2}{*}{J.}
& \multicolumn{2}{c}{OpenVLA-OFT}
& \multicolumn{3}{c}{$\pi_{0.5}$} \\
\cmidrule(lr){3-4}\cmidrule(lr){5-7}
& & Base & J-PARC
& Base & RVLA & J-PARC \\
\midrule
\midrule
1.0 & - & \textbf{95.8} & 95.4 & 96.6 & 96.2 & \textbf{97.0}\\
\midrule

\multirow{8}{*}{150}
& $j_0$ & 59.9 & \textbf{68.7} & 69.4 & 65.3 & \textbf{75.6} \\
& $j_1$ & \textbf{51.3} & 50.1 & 64.3 & 58.5 & \textbf{65.8} \\
& $j_2$ & 31.8 & \textbf{41.7} & 46.6 & 43.3 & \textbf{63.0} \\
& $j_3$ & 5.9 & \textbf{6.3} & 9.6 & 10.0 & \textbf{11.8} \\
& $j_4$ & 80.2 & \textbf{82.0} & 84.6 & 84.2 & \textbf{86.1} \\
& $j_5$ & 78.9 & \textbf{83.7} & 87.9 & 88.1 & \textbf{91.9} \\
& $j_6$ & 58.9 & \textbf{65.3} & 67.4 & 66.6 & \textbf{72.2} \\
\cmidrule(lr){2-7}
& Avg. & 52.4 & \textbf{56.8} & 61.4 & 59.4 & \textbf{66.6} \\

\midrule

\multirow{8}{*}{200}
& $j_0$ & 51.7 & \textbf{63.7} & 62.1 & 61.3 & \textbf{72.0} \\
& $j_1$ & 35.5 & \textbf{35.6} & 51.8 & 49.2 & \textbf{54.0} \\
& $j_2$ & 30.9 & \textbf{37.5} & 43.7 & 40.1 & \textbf{57.6} \\
& $j_3$ & \textbf{5.2} & 4.6 & 7.1 & 5.9 & \textbf{7.5} \\
& $j_4$ & 80.0 & \textbf{81.8} & 84.9 & 84.2 & \textbf{85.9} \\
& $j_5$ & 79.2 & \textbf{83.0} & 88.3 & 87.9 & \textbf{92.1} \\
& $j_6$ & 59.0 & \textbf{64.9} & 66.3 & 66.8 & \textbf{72.0} \\
\cmidrule(lr){2-7}
& Avg. & 48.8 & \textbf{53.0} & 57.7 & 56.5 & \textbf{63.0} \\

\midrule

\multirow{8}{*}{300}
& $j_0$ & 50.7 & \textbf{59.9} & 60.5 & 58.7 & \textbf{69.2} \\
& $j_1$ & 20.2 & \textbf{20.5} & \textbf{34.0} & 29.4 & 32.2 \\
& $j_2$ & 30.3 & \textbf{36.4} & 44.0 & 39.6 & \textbf{55.6} \\
& $j_3$ & \textbf{5.1} & 5.0 & 5.9 & 5.8 & \textbf{6.6} \\
& $j_4$ & 80.2 & \textbf{82.2} & 84.3 & 83.2 & \textbf{86.0} \\
& $j_5$ & 79.2 & \textbf{83.7} & 89.0 & 87.4 & \textbf{92.5} \\
& $j_6$ & 59.0 & \textbf{65.2} & 67.7 & 66.1 & \textbf{72.3} \\
\cmidrule(lr){2-7}
& Avg. & 46.4 & \textbf{50.4} & 55.1 & 52.9 & \textbf{59.2} \\

\bottomrule
\end{tabular}
\end{adjustbox}
\end{minipage}

\endgroup
\end{table}
\footnotetext{\url{https://github.com/gakakulicc/RobustVLA}}
We use LIBERO~\cite{libero} benchmark to evaluate whether J-PARC improves the closed-loop robustness of frozen VLA policies under joint-level physical fault. 
For efficient real-world data collection, we collect residual targets and train J-PARC only under the locked-joint setting, avoiding the need to gather separate calibration data for each friction or range-limit condition.
Our experiments are designed to answer three questions. 
First, does J-PARC improve robustness under both seen and unseen joint faults while preserving the behavior of the base policy in fault-free environments? 
Second, does joint fault regime encoder perceive faulty joint under both seen and unseen joint fault?
Third, does J-PARC generalize to real-world environment?

We instantiate J-PARC on top of two frozen VLA backbones, OpenVLA-OFT and $\pi_{0.5}$, and compare each policy with and without J-PARC. 
All methods are evaluated by task success rate. 
J-PARC is applied only to the translational arm action dimensions, while the base VLA policy, rotation command, and gripper command remain unchanged. 
This evaluation therefore tests whether a lightweight residual module can compensate for physical execution mismatch without fine-tuning the VLA backbone itself. Detailed experiment setting is declared in Appendix~\ref{app:eval_setup}.

\subsection{Main Results}

Table~\ref{tab:main_results_single_joint_filtered} shows that J-PARC improves robustness across both VLA backbones while preserving performance on fault-free environment. Under joint lock conditions, J-PARC improves the average success rate from 42.3\% to 46.8\% on OpenVLA-OFT and from 48.4\% to 55.2\% on $\pi_{0.5}$. 

Under increased joint friction, which is not observed during training, J-PARC also improves the average success rate, demonstrating its ability to generalize to unseen friction-induced execution mismatch without collecting additional friction-specific rollouts. Moreover, J-PARC leaves the base policy behavior nearly unchanged in fault-free environments indicating that J-PARC does not substantially interfere with the base policy under nominal robot dynamics.


\footnotetext[\value{footnote}]{\url{https://github.com/gakakulicc/RobustVLA}}

\begin{wrapfigure}{t}{0.65\textwidth}
    \centering
    \vspace{-1.8em}
    \includegraphics[width=\linewidth]{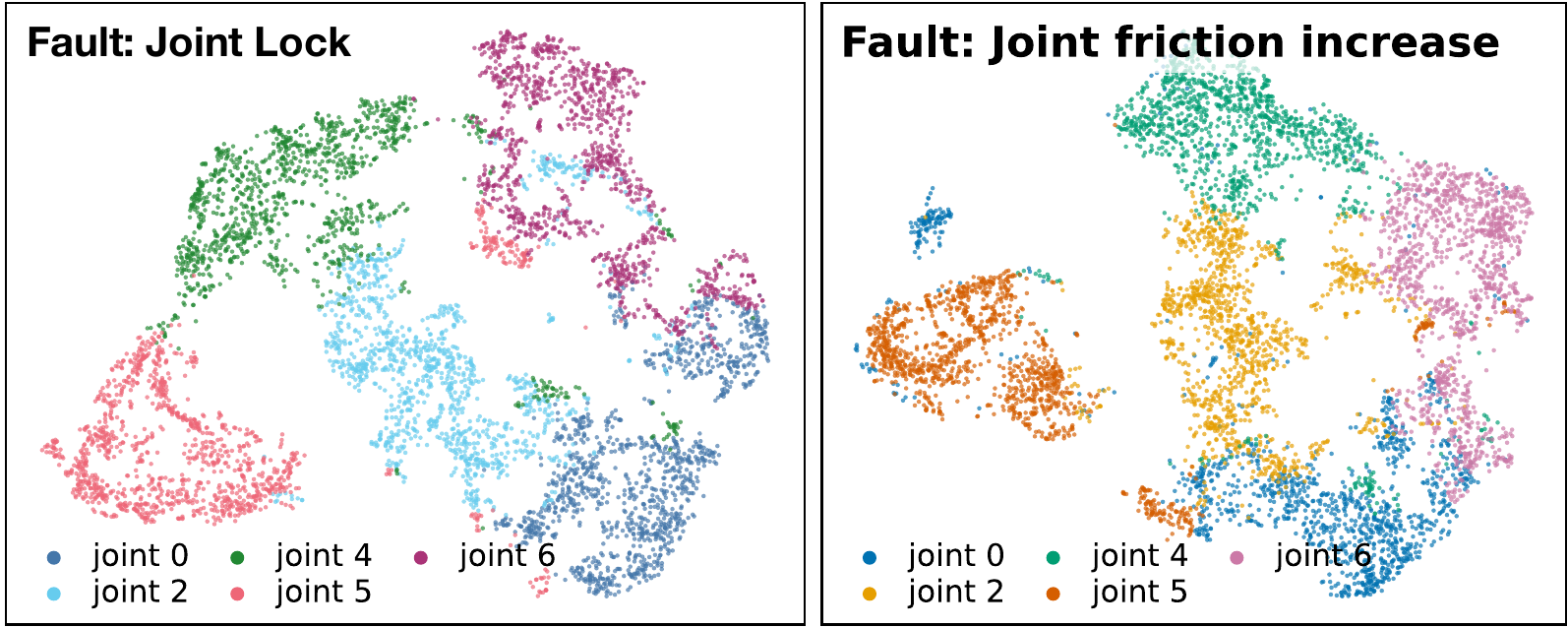}
    \vspace{-0.8em}
    \caption{
    \textbf{T-SNE visualization of latent representations learned by the joint-fault regime encoder.}
    We visualize encoder embeddings under locked-joint faults and increased-friction faults.
    The embeddings form joint-dependent clusters under both fault types.
    }
    \label{fig:joint_regime}
    \vspace{-1.5em}
\end{wrapfigure}
\subsection{Effectiveness of the Joint Fault Regime Encoder}
To evaluate whether the joint-fault regime encoder captures meaningful execution contexts, we visualize its latent representations under different joint-level fault conditions.
As shown in Fig.~\ref{fig:joint_regime}, the learned embeddings form distinct clusters according to the affected joint under both locked-joint and increased-friction faults.
This indicates that recent robot states and base VLA actions contain sufficient information to infer the current joint-fault regime.
Importantly, the embedding structure differs across fault types, suggesting that the encoder does not merely identify the joint index, but also captures how each fault condition changes the robot's execution dynamics.
These results support the design of J-PARC, where the residual calibrator conditions action correction on the inferred joint-fault regime.
\FloatBarrier
\subsection{Real-World Validation}
\label{sec:real_world}

We further validate Joint-level fault and J-PARC on a real-world setting with Trossen WidowX AI robot using a $\pi_{0.5}$ policy for a bowl pick-and-place task. 
We evaluate fault-free execution and two persistent joint-level fault conditions, where joint 4 or joint 5 is held locked during closed-loop execution. 
We focus on joints 4 and 5 because faults on the other joints often make the task nearly infeasible on the real robot or can impose risk on the hardware during repeated evaluation. 
The real-world calibrator is trained from fault-free reference rollouts and teacher-oracle residual targets that correct the faulted rollout toward the corresponding fault-free end-effector trajectory.

\begin{figure*}[t]
\centering
\includegraphics[width=0.9\textwidth]{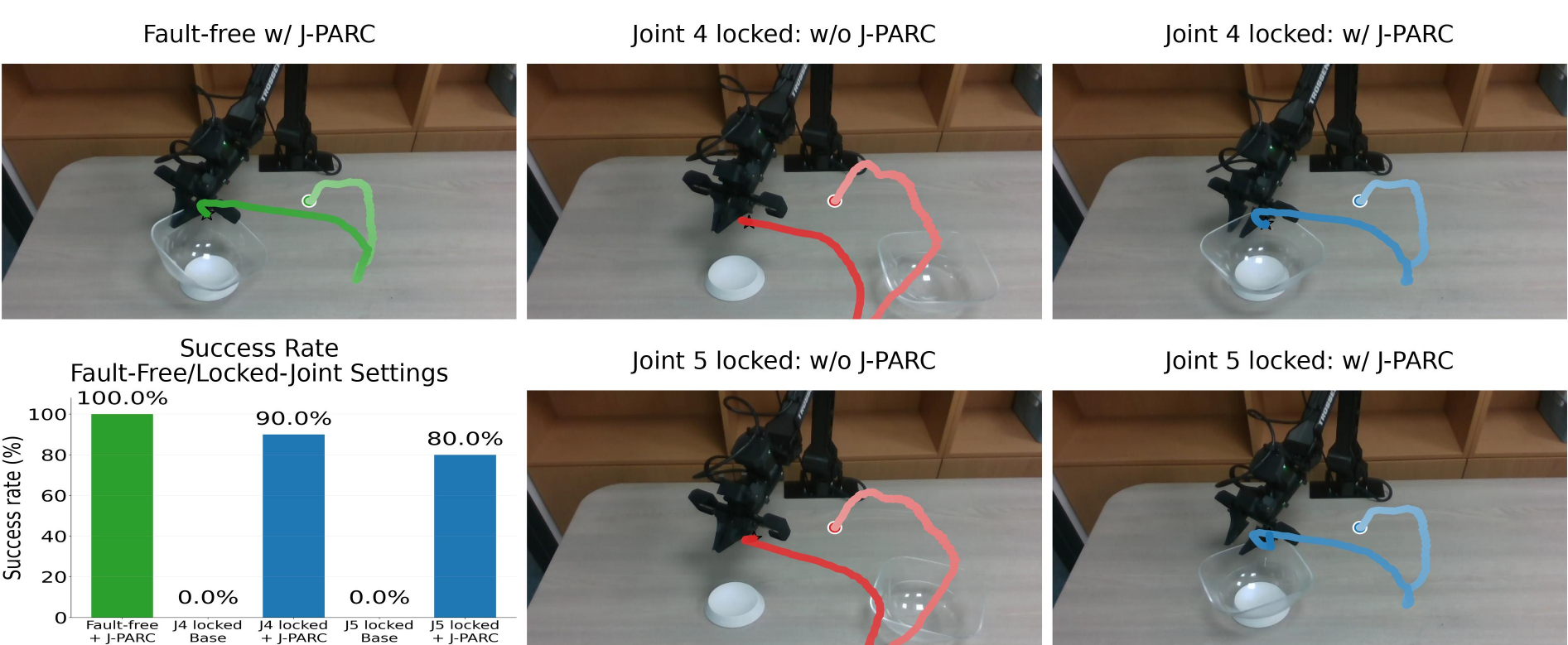}
\caption{
\textbf{Real-world evaluation on the Trossen WidowX AI bowl pick-and-place task.} 
Trajectory overlays are drawn on the final observation frame. 
Under joint-level faults, the base policy without J-PARC often drifts away from the fault-free execution path, while J-PARC redirects the end-effector trajectory toward the successful placement behavior.
}
\label{fig:real_world_jparc}
\end{figure*}

Fig.~\ref{fig:real_world_jparc} summarizes real-world validations. 
J-PARC preserves performance in fault-free environment, succeeding in all trials.
Under joint lock, the base policy fails in all trials, while J-PARC succeeds in most trials. The trajectory visualization shows that joint-level faults induce a persistent execution mismatch that pushes the robot away from the nominal rollout, while the residual correction keeps the end-effector closer to the fault-free reference path without changing the frozen VLA backbone.

\section{Related Work}

\paragraph{Robustness of Vision-Language-Action Models.}
Recent VLA models \cite{openpi05,openvlaoft,rt1,oxe,octo,rt2, vlasurvey2, groot, openpifast, smolvla, vlasurvey1} have enabled generalist robotic manipulation policies by mapping visual observations and language instructions to actions. As these models have become more capable, recent work~\cite{liberoplus, robustvla, vlatest, rovla2, patch} has begun to evaluate their robustness under distribution shifts and perturbations \cite{robustvla}. Benchmarks such as LIBERO-Plus \cite{liberoplus} and VLATest \cite{vlatest} study robustness to changes in object layout, visual appearance, camera viewpoint, background, lighting, sensor noise, and language instructions. Other studies \cite{evavla, patch, vlaattack, badvla} investigate adversarial visual perturbations, including patches or image-space attacks, and show that perception-side corruption can significantly degrade closed-loop manipulation performance. These works provide important tools for stress-testing VLA policies, but they primarily focus on perturbations to the observation or task specification rather than perturbations that arise from the robot body during action execution.
\paragraph{Action-Space Robustness and Physical Execution Mismatch.}
Beyond visual perturbations, recent studies have examined whether VLA policies are robust to perturbations in the action space \cite{robustvla}. RobustVLA evaluates VLA models under multi-modal perturbations, including action, observation, instruction, and environment perturbations, and shows that action-space perturbations can be particularly harmful \cite{robustvla}. Adversarial vulnerability studies of VLA-based robotic systems further demonstrate that attacks can destabilize actions, manipulate trajectories, or degrade task execution through adversarial objectives \cite{patch}. However, these approaches mainly perturb the commanded action or the sensory input. Joint-level physical faults represent a different failure mode: the action command may be unchanged, but the robot body realizes it differently because a joint is locked, range-limited, or affected by increased friction \cite{joint2, joint1, jointsurvey}.

\section{Limitations}

For hardware safety, We focus on joint 4 and 5, and our real-world validation is limited to a bowl pick-and-place task on the Trossen WidowX AI robot. Therefore, a broader evaluation across diverse robots, tasks, and fault locations remains future work.

In addition, J-PARC currently relies on fault-free reference rollouts and teacher-oracle residual targets, which assume access to task-specific reference trajectories.
Future work should explore automatic reference construction, self-supervised residual learning, online adaptation, and calibration over richer action spaces beyond translational Cartesian actions.

\section{Conclusion}


We studied joint-level fault conditions as a realistic robustness challenge for deploying VLA policies. Joint faults alter the way actions are physically executed, creating a mismatch between commanded actions and robot-realized motions.
Our analysis shows that these faults have heterogeneous effects across joints, can degrade performance even when tasks remain feasible, and may accumulate into closed-loop execution mismatch.

To address this challenge, we proposed J-PARC, a lightweight residual calibration framework.
J-PARC infers the current joint-fault regime from recent execution history and corrects actions to keep faulted rollouts closer to the nominal trajectory expected by the base policy.
Experiments in simulation and real-world settings show that J-PARC improves robustness under joint-level faults while preserving fault-free behavior.
These results highlight fault-aware action calibration as a practical step toward reliable VLA deployment under physical fault conditions.

\newpage
\bibliography{example}  
\newpage
\setcounter{table}{0}

\appendix

\section{Implementation and Evaluation Details}
\label{app:eval_setup}

\subsection{Implementation Details}
\label{app:implementation_details}

We summarize the implementation using the notation introduced in Sec.~\ref{sec:method}.
Table~\ref{tab:jparc_notation_impl} maps the main symbols in Sec.~\ref{sec:jparc_overview} to the corresponding J-PARC inputs.

\begin{table}[H]
\centering
\small
\caption{
\textbf{J-PARC notation and token mapping.}
}
\label{tab:jparc_notation_impl}
\begin{tabular}{@{}p{0.32\linewidth}p{0.62\linewidth}@{}}
\toprule
Symbol & Implementation role \\
\midrule
$A_t^{\mathrm{base}}$ & Base action chunk, length $K=8$ \\
$a_{t,k}^{\mathrm{base}}$ & 7-D action: xyz, rotation, gripper \\
$c_t=[s_t,q_t,\dot q_t]$ & Current token; $q_t$: joint position, $\dot q_t$: joint velocity \\
$\mathcal{H}_t=\{h_i\}_{i=t-H}^{t-1}$ & History buffer, length $H=6$ \\
$h_i=[a_i^{\mathrm{base}},c_i,\delta_i^{\mathrm{hist}}]$ & History token \\
$\Delta q_{t-H:t}$ & Joint-displacement window \\
$u_t=g_\psi(\Delta q_{t-H:t})$ & Joint-regime token \\
$R_{\theta,t}$ & Residual chunk, $K\times3$ \\
$x_i,x_i^{\mathrm{ref}}$ & Faulted / reference EE position \\
$\tau_{\mathrm{demo}}$ & threshold for teacher correction \\
$r_{t,k}^{\star}$ & Teacher residual target \\
$\tilde a_{t,k}$ & Corrected action \\
\bottomrule
\end{tabular}
\end{table}

The joint-regime encoder $g_\psi$ is trained before the residual calibrator.
It uses recent joint displacement $\Delta q_{t-H:t}$ as input, embeds each per-step displacement with an MLP, and processes the sequence with a GRU to produce $u_t$.
For residual training, we initialize from this pretrained $g_\psi$ and keep it frozen.

The residual calibrator $f_\theta$ is a transformer encoder over the current token $c_t$, history tokens $h_i$, base-action tokens $A_t^{\mathrm{base}}$, and the joint-regime token $u_t$.
It predicts $R_{\theta,t}=\{r_{\theta,t,k}\}_{k=0}^{K-1}$ and applies the correction following Eq.~\ref{eq:corrected_action}.
The rotation and gripper commands are preserved as in Sec.~\ref{sec:jparc_overview}.
Table~\ref{tab:jparc_impl_hparams} summarizes the two-stage training setup, where $K$ denotes the action-chunk horizon and $H$ denotes the execution-history length.

\begin{table}[H]
\centering
\scriptsize
\caption{
\textbf{Two-stage J-PARC implementation hyperparameters.}
}
\label{tab:jparc_impl_hparams}
\begin{tabular}{@{}p{0.18\linewidth}p{0.38\linewidth}p{0.38\linewidth}@{}}
\toprule
Stage & OpenVLA-OFT & $\pi_{0.5}$ \\
\midrule
Common &
\begin{tabular}[t]{@{}ll@{}}
Train joints & $j_0,j_2,j_4,j_5,j_6$ \\
Excluded joints & $j_1,j_3$ \\
Reason & feasibility/data risk
\end{tabular} &
\begin{tabular}[t]{@{}ll@{}}
Train joints & $j_0,j_2,j_4,j_5,j_6$ \\
Excluded joints & $j_1,j_3$ \\
Reason & feasibility/data risk
\end{tabular} \\
\midrule
Regime pretrain &
\begin{tabular}[t]{@{}ll@{}}
Input & $\Delta q_{t-H:t}$ \\
MLP & hidden 64, embed 128 \\
GRU & hidden 128 \\
Latent & $d_u=128$ \\
Train & 100 ep, batch 128 \\
Loss & BCE + fault-joint CE
\end{tabular} &
\begin{tabular}[t]{@{}ll@{}}
Input & $\Delta q_{t-H:t}$ \\
MLP & hidden 64, embed 128 \\
GRU & hidden 128 \\
Latent & $d_u=128$ \\
Train & 100 ep, batch 128 \\
Loss & BCE + fault-joint CE
\end{tabular} \\
\midrule
Residual training &
\begin{tabular}[t]{@{}ll@{}}
Init & frozen $g_\psi$ \\
Data & successful trajectory\\
& on joint lock environment \\
$\tau_{\mathrm{demo}}$ & 0.02 m \\
Shape & $K=8$ actions, $H=6$ history steps \\
Model & Transformer, $d=256$ \\
Arch & 4 layers, 8 heads, feed-forward 512 \\
Train & 40 ep, batch 16 
\end{tabular} &
\begin{tabular}[t]{@{}ll@{}}
Init & frozen $g_\psi$ \\
Data & successful trajectory \\
& on joint lock environment \\
$\tau_{\mathrm{demo}}$ & 0.02 m \\
Shape & $K=8$ actions, $H=6$ history steps \\
Model & Transformer, $d=256$ \\
Arch & 4 layers, 8 heads, feed-forward 512 \\
Train & 40 ep, batch 128 
\end{tabular} \\
\bottomrule
\end{tabular}
\end{table}

We exclude joints $j_1$ and $j_3$ from residual supervision because preliminary feasibility checks showed that these faults often lead to near-infeasible or unstable executions.
Including such rollouts as supervision can contaminate the residual dataset with targets that reflect unrecoverable failures rather than correctable command--execution mismatch.
We therefore use $j_1$ and $j_3$ as evaluation conditions where applicable, but do not use them to train the residual calibrator.

\subsection{Evaluation Setup}

Table~\ref{tab:main_results_single_joint_filtered} evaluates two joint-level fault families: joint limits and increased joint friction.
Both panels include a fault-free row, denoted by range or scale $1.0$, and report success rates averaged over the LIBERO-Spatial, LIBERO-Object, LIBERO-Goal, and LIBERO-10 suites.
Each suite contains 10 tasks, and we run 50 episodes per task, resulting in 500 episodes per suite and condition.
All policies execute 8-step action chunks, and all perturbation runs use seed 0.
For OpenVLA-OFT, we evaluate the suite-specific LIBERO-finetuned model without 8-bit or 4-bit quantization.
For $\pi_{0.5}$, we use the same 8-step replanning period in the OpenPI LIBERO evaluator.

\begin{table}[H]
\centering
\small
\caption{
\textbf{Evaluation protocol for Table~\ref{tab:main_results_single_joint_filtered}.}
}
\label{tab:table2_eval_hparams}
\begin{tabular}{@{}p{0.30\linewidth}p{0.64\linewidth}@{}}
\toprule
Item & Setting \\
\midrule
Suites & Spatial, Object, Goal, 10 \\
Episodes & 50/task, 500/suite \\
Metric & Mean suite success rate \\
Execution & 8-step chunks \\
Joint-limit panel & Range $1.0$, $0.05$, $0.03$, Locked \\
Friction panel & Scale $1.0$, $150$, $200$, $300$ \\
Affected joints & $j_0,\ldots,j_6$ columns \\
\bottomrule
\end{tabular}
\end{table}

In the joint-limit panel, the range column specifies how strongly the selected joint is constrained.
Range $1.0$ is the fault-free setting.
For range-limited rows, the target joint range is replaced by a narrow interval centered at the current joint position; if the original joint span is $s$ and the range ratio is $r$, the interval half-width is $0.5sr$, clipped to the original joint limits.
The Locked row freezes the target joint for the entire episode.

In the friction panel, scale $1.0$ is the fault-free setting.
For scales $150$, $200$, and $300$, we enable plant degradation on the target joint with damping scale $1.0$, damping offset $0.0$, friction-loss scale equal to the table value, friction-loss offset $0.0$, and actuator force-range scale $1.0$.

For J-PARC, we use one suite-specific residual calibrator checkpoint per backbone and LIBERO suite, trained as summarized in Table~\ref{tab:jparc_impl_hparams}.
The residual calibrator predicts $R_{\theta,t}$ for $a_{t,k}^{\mathrm{base},xyz}$ and preserves rotation and gripper commands.
At evaluation time, $r_{\theta,t,k}$ is added to $a_{t,k}^{\mathrm{base},xyz}$ and the corrected action is clipped to $[-1,1]$.
CIK$^\ast$ is evaluated only in the joint-limit panel because it assumes oracle identification of the constrained joint; it is omitted from the friction panel because CIK cannot be directly applied in friction-fault environments.



\section{Usage-Weighted End-Effector Sensitivity}
\label{app:sensitivity}

Let $\mathcal{D}_{\mathrm{fault\text{-}free}}^{\pi}$ denote the set of successful fault-free rollouts generated by the base policy $\pi$.
Each rollout is written as $\tau = (q_1,\ldots,q_{T_\tau})$, where $q_t \in \mathbb{R}^{7}$ is the robot joint configuration at timestep $t$.
For a gripper probe site $s \in \mathcal{G}$, let $p_s(q_t) \in \mathbb{R}^{3}$ denote its Cartesian position and let
$J_s(q_t) \in \mathbb{R}^{3 \times 7}$ denote the translational Jacobian:
\[
J_s(q_t)
=
\frac{\partial p_s(q_t)}{\partial q_t}.
\]

We first define the task-specific usage of joint $j$ in a rollout $\tau$ as the normalized joint-range variation:
\[
u_j(\tau)
=
\frac{
\max_{1 \le t \le T_\tau} q_{t,j}
-
\min_{1 \le t \le T_\tau} q_{t,j}
}{
q^{\max}_j - q^{\min}_j
}.
\]
The average joint usage over successful fault-free rollouts is
\[
\bar{u}_j
=
\frac{1}{|\mathcal{D}_{\mathrm{fault\text{-}free}}^{\pi}|}
\sum_{\tau \in \mathcal{D}_{\mathrm{fault\text{-}free}}^{\pi}}
u_j(\tau).
\]

We then define the average end-effector sensitivity of joint $j$ as the rollout-averaged maximum translational Jacobian norm over gripper probe sites:
\[
\bar{\kappa}_j
=
\frac{1}{|\mathcal{D}_{\mathrm{fault\text{-}free}}^{\pi}|}
\sum_{\tau \in \mathcal{D}_{\mathrm{fault\text{-}free}}^{\pi}}
\frac{1}{T_\tau}
\sum_{t=1}^{T_\tau}
\max_{s \in \mathcal{G}}
\left\|
\left[J_s(q_t)\right]_{:,j}
\right\|_2 .
\]

Finally, the usage-weighted single-joint sensitivity is defined as
\[
S_{\mathrm{single}}^{\mathrm{usage}}(j)
=
\bar{u}_j \, \bar{\kappa}_j .
\]
For visualization, we normalize the sensitivity across joints:
\[
\widetilde{S}_{\mathrm{single}}^{\mathrm{usage}}(j)
=
\frac{
S_{\mathrm{single}}^{\mathrm{usage}}(j)
}{
\max_{k \in \{0,\ldots,6\}}
S_{\mathrm{single}}^{\mathrm{usage}}(k)
}.
\]

\section{Ablation Study of J-PARC components}
\begin{table}[H]
\centering
\caption{
\textbf{Ablation of J-PARC components on $\pi_{0.5}$.}
Success rates (\%) are reported under locked-joint faults.
Joint Regime indicates the use of the joint-regime representation, and History indicates the use of history token.
}
\label{tab:openpi_jparc_ablation}
\begin{tabular}{lcc|ccccc|c}
\toprule
Method & Joint Regime & History & $j_0$ & $j_2$ & $j_4$ & $j_5$ & $j_6$ & Avg. \\
\midrule
Base $\pi_{0.5}$ & -- & -- & 58.8 & 38.8 & 84.4 & 87.1 & 64.5 & 66.7 \\
J-PARC & \xmark & \xmark & 65.6 & 55.1 & 82.4 & 81.7 & 70.7 & 71.1 \\
J-PARC & \cmark & \xmark & 68.0 & \textbf{57.3} & 83.8 & 82.1 & 70.4 & 72.3 \\
J-PARC & \cmark & \cmark & \textbf{72.2} & 56.1 & \textbf{87.6} & \textbf{91.2} & \textbf{73.8} & \textbf{76.2} \\
\bottomrule
\end{tabular}
\end{table}
Table~\ref{tab:openpi_jparc_ablation} shows the ablation of J-PARC components on $\pi_{0.5}$ under locked-joint faults.
Even without the joint-regime representation or history, J-PARC improves the average success rate over the base policy from \(66.7\%\) to \(71.1\%\), showing the benefit of residual calibration itself.
Adding the joint-regime representation further improves the average success rate to \(72.3\%\), while incorporating execution history achieves the best average performance of \(74.3\%\).
These results indicate that J-PARC benefits from residual correction, fault-regime awareness, and history-based modeling of command--execution mismatch.

\end{document}